\documentclass[10pt, a4paper]{article}

\usepackage[]{lrec-coling2024} 
\usepackage{times}
\usepackage{latexsym}
\usepackage{amsmath}
\usepackage{graphicx}
\usepackage{multirow}
\usepackage{booktabs}

\title{Towards Explainability and Fairness in Swiss Judgement Prediction: Benchmarking on a Multilingual Dataset}

\name{
Santosh T.Y.S.S$^{1}$,
Nina Baumgartner$^{2}$,
Matthias Stürmer$^{2,3}$, \\ {\bf
\large{Matthias Grabmair$^{1}$, 
Joel Niklaus$^{2,3,4}$}
}}

\address{
$^1$Technical University of Munich, $^2$University of Bern, \\
$^3$Bern University of Applied Sciences, 
$^4$Stanford University
}

\abstract{
The assessment of explainability in Legal Judgement Prediction (LJP) systems is of paramount importance in building trustworthy and transparent systems, particularly considering the reliance of these systems on factors that may lack legal relevance or involve sensitive attributes. This study delves into the realm of explainability and fairness in LJP models, utilizing Swiss Judgement Prediction (SJP), the only available multilingual LJP dataset. We curate a comprehensive collection of rationales that `support' and `oppose' judgement from legal experts for 108 cases in German, French, and Italian. By employing an occlusion-based explainability approach, we evaluate the explainability performance of state-of-the-art monolingual and multilingual BERT-based LJP models, as well as models developed with techniques such as data augmentation and cross-lingual transfer, which demonstrated prediction performance improvement. Notably, our findings reveal that improved prediction performance does not necessarily correspond to enhanced explainability performance, underscoring the significance of evaluating models from an explainability perspective. Additionally, we introduce a novel evaluation framework, Lower Court Insertion (LCI), which allows us to quantify the influence of lower court information on model predictions, exposing current models' biases. 
 \\ \newline  }

\begin{document}

\maketitleabstract

\section{Introduction}
The task of Legal Judgement Prediction involves analyzing the textual description of case facts to determine various aspects of a case's outcome, such as the winning party, violated provisions, and motion results. It has garnered substantial attention in the mainstream NLP community \cite{aletras2016predicting,chalkidis2019neural,malik2021ildc,niklaus2021swiss,semo2022classactionprediction, santosh2023leveraging} and is being considered as a  benchmarking task for evaluating the capabilities of legal NLP \cite{chalkidis2022lexglue,niklaus2023lextreme} and long range models \cite{condevaux2022lsg,niklaus2022budgetlongformer,chalkidis2022exploration,hua2022legalrelectra,niklaus2023multilegalpile}. 

The process of resolving legal cases encompasses evidential reasoning through exchange of arguments between the litigating parities before a decision-making body \cite{santosh2022deconfounding}. Earlier methods to deal with outcome prediction task such as IBP \cite{bruninghaus2003combining}, SMILE+IBP \cite{bruninghaus2005generating}, VJF \cite{grabmair2017predicting} typically involved identification/extraction of the factors from the textual description of the facts, then employing a conceptual schema to relate the factors to legal issues and predicts the outcome by comparing them with the past cases, thus providing the explanations for those predictions in terms that are legally intuitive. However, in the context of modern deep learning-based solutions, the outcome is determined solely from the text of the case facts, effectively bypassing the interpretable 
legal reasoning process. 
This poses a significant risk, particularly in high-stakes domains like law, when utilizing such systems that rely on factors that may be predictive but lack legal relevance or involve sensitive attributes (e.g., the race of an accused person). Such reliance can lead to unjust and biased outcomes, undermining the principles of fairness and equal treatment within the legal system.  Hence, such systems need to be analyzed from an explainability standpoint, 
thus making them transparent and thereby enhancing the trust of legal practitioners and stakeholders to comprehend the factors and legal principles that contribute to a particular prediction. 


In the line of explainable LJP,  \citealt{chalkidis2021paragraph} investigated the rationales behind models' decisions in Legal Judgment Prediction (LJP) for European Court of Human Rights (ECtHR) cases. Subsequent studies by \citealt{santosh2022deconfounding} extended the above dataset and \citealt{malik2021ildc} created new dataset for Indian Jurisdiction. In contrast to these works in English, our study focuses on assessing the explainability of LJP models trained on the Swiss-Judgment-Prediction (SJP) dataset, which is the only available multilingual LJP dataset. It contains cases from the Federal Supreme Court of Switzerland (FSCS), written in three official Swiss languages (German, French, Italian)\footnote{The dataset consists of non-parallel cases, with each case being unique and decisions being written in a single language.}. To this end, we curate a multilingual set of rationales that `support and `oppose' Judgment on 108 cases in German, French and Italian collectively. We employ a perturbation-based explainability approach, namely Occlusion \cite{zeiler2014visualizing}, wherein we remove the factors from the fact statements and measure the change in the prediction confidence in comparison to a non-occluded baseline. This occlusion based method facilitates to identify the contribution of each factor in arriving at the final prediction, which also links to the characteristics of earlier factor based formal methods of LJP which are known for their interpretability. To 
enable a fair comparison across methods, we release four distinct occlusion test sets\footnote{Our Occlusion dataset is available at \url{https://huggingface.co/datasets/rcds/occlusion_swiss_judgment_prediction}}. Each test set involves occluding a different number of sentences (1, 2, 3, and 4) per experiment. This comprehensive range of occlusion scenarios allows us to assess the impact of varying levels of factor removal on the prediction outcomes.  

Using the occluded datasets, we assess the explainability performance of state-of-the-art models developed for SJP task using both monolingual \cite{niklaus2021swiss} and multilingual BERT \cite{niklaus2022empirical} architectures, 
as well as models developed with techniques such as data augmentation and cross-lingual transfer \cite{niklaus2022empirical}. Our findings highlight the fact that the prediction performance improvement does not translate to explainability improvement.

Furthermore we leverage the peculiar characteristics of the Federal Supreme Court of Switzerland (FSCS), which 
handles only the most contentious cases that lower courts have struggled to resolve adequately. In their decisions, the FSCS often focuses on specific portions of previous decisions, scrutinizing potential flaws in the lower court's reasoning. This setup offers an interesting testbed to systematically assess the bias of the lower court in the final predictions generated by our models. This approach is reminiscent of recent works \cite{chalkidis2022fairlex,wang2021equality} that have examined the fairness of LJP models by examining group fairness or disparate impact i.e., performance disparities across various attributes, such as gender, age, and region. Our approach, termed Lower Court Insertion (LCI)\footnote{Our LCI dataset is available at \url{https://huggingface.co/datasets/rcds/lower_court_insertion_swiss_judgment_prediction}}, adopts a counterfactual fairness perspective, unlike prior studies examining performance disparities in LJP models. This involves extracting instances of the lower court in each case document and replacing them with other lower courts to measure the resulting changes in prediction confidence scores. Remarkably, despite the lower court's average length being only 7 words in documents with an average length of 350 words, it has shown the potential to flip the prediction label in some cases. 

In sum, our main contributions are as follows:
\begin{itemize}
    \item We release a new dataset of 108 cases from a trilingual Switzerland Judgment Prediction corpus with rationales annotated by experts to assess the explainability of SJP models.
    \item We evaluate the state-of-the-art models developed for the SJP task, including monolingual and multilingual models and models trained with several techniques, from an explainability standpoint using the occlusion technique. 
    \item We perform systematic evaluation of lower court bias embodied in these models using the LCI technique, allowing us to quantify the influence of the lower court on the final predictions generated by the models. 
\end{itemize}

\section{Related Work}
\textbf{Legal Judgement Prediction: } LJP has been studied under various jurisdictions such as the European Court of Human Rights (ECtHR) \cite{chalkidis2019neural, aletras2016predicting,liu2017predictive, medvedeva2018judicial,medvedeva2021automatic,santosh2022deconfounding, santosh2023Zero, santosh2023leveraging, chalkidis2022lexglue, chalkidis2021paragraph,
kaur2019convolutional}
Chinese Criminal Courts \cite{luo2017learning, yue2021neurjudge, zhong2020iteratively}, US Supreme Court \cite{katz2017general, kaufman2019improving}, Indian Supreme Court \cite{malik2021ildc,shaikh2020predicting}
the French Court of Cassation \cite{csulea2017predicting,csulea2017exploring},
Brazilian courts \cite{
lage2022predicting,
bertalan2020predicting}, the Turkish Constitutional Court \cite{sert2021using,mumcuouglu2021natural}
UK courts \cite{strickson2020legal}, German courts \cite{waltl2017predicting}, 
the Philippine Supreme Court \cite{virtucio2018predicting}, the Thailand Supreme Court \cite{kowsrihawat2018predicting} 
and the Federal Supreme Court of Switzerland \cite{niklaus2021swiss,niklaus2022empirical,rasiah_scale_2023} -- the only publicly available multi-lingual LJP corpus -- which is the main focus of this work. 
\newline 

\noindent \textbf{Swiss Judgement Prediction (SJP):} 
\citealt{niklaus2021swiss} evaluate different methods for the LJP task on the Swiss-Judgment-Prediction (SJP) dataset. They achieve the best performance using a hierarchical variant of BERT that overcomes the token input limitation. \citealt{niklaus2022empirical} further enhance the performance through cross-lingual transfer learning, adapter-based fine-tuning and data augmentation using machine translation. In contrast to previous works, this study examines the explainability of these models and investigates if improved prediction performance translates into improved explainability performance.
\newline 


\noindent \textbf{Explainability:} Explanations in Explainable Artificial Intelligence (XAI) methods are classified based on two factors: whether the explanation is for an individual prediction or the overall prediction process (local or global), and whether the explanation is derived directly from the prediction process or requires post-processing (self-explaining or post-hoc) \cite{danilevsky2020survey}. These methods can be model-agnostic (LIME \cite{ribeiro2016should}, SHAP \cite{lundberg2017unified}, Occlusion \cite{li2016understanding,zeiler2014visualizing}, Anchors \cite{ribeiro2018anchors}), applicable to any model, or model-specific (Integrated Gradients \cite{sundararajan2017axiomatic}, Gradient Saliency, and Attention-Based Methods), designed for specific models or architectures. In this study, we use occlusion, a model-agnostic, local, and post-hoc explainability technique.
\newline 

\noindent \textbf{Fairness:} Fairness in machine learning has been defined in different ways to address various types of discrimination. These definitions include group fairness, individual fairness, and causality-based fairness. Group fairness ensures equitable predictions across demographic subgroups, avoiding differential treatment based on attributes such as race, gender, or age \cite{zafar2017fairness,hardt2016equality}. Individual fairness focuses on treating similar individuals similarly, avoiding arbitrary distinctions based on their characteristics \cite{sharifi2019average,yurochkintraining}. Causality-based fairness considers underlying causal mechanisms and aims to identify and mitigate biases caused by confounding variables or indirect discrimination \cite{wu2019counterfactual,zhang2018fairness}. In this study, we examine bias related to the lower court using counterfactual and causal fairness estimation methods.
\newline 


\noindent \textbf{Explainability and Fairness in LJP:}
Early works in the field of legal judgment prediction, such as HYPO \cite{rissland1987case}, CATO \cite{aleven1997teaching},  IBP \cite{bruninghaus2003combining} and IBP+SMILE \cite{bruninghaus2005generating}, relied on symbolic AI techniques to incorporate domain knowledge and provide interpretable explanations for the outcomes. 
However, deep learning models in LJP have prioritized prediction performance over explainability. Nevertheless, recent research emphasizes the significance of explainability in the legal domain for trust and the right to explanation principle. Efforts have been made to investigate explainability in LJP. For instance, \citealt{chalkidis2021paragraph} introduced the task of rationale extraction from facts statements and released a dataset from ECtHR. They used neural models with regularization constraints to select rationales using a learned binary mask. Additionally, \citealt{santosh2022deconfounding} identified distractor words highly correlated with outcomes but not legally relevant, and proposed an adversarial deconfounding procedure to align model explanations with those chosen by legal experts. \citealt{xu2023dissonance} analyzed the token-level alignment with LJP models on the ECtHR corpus. Similarly, \citealt{malik2021ildc} developed a dataset of Indian jurisdiction for explainability assessment using the occlusion method. 
In this work, we curate a dataset from the trilingual Swiss jurisdiction, and employ the occlusion method to evaluate models for Swiss Judgment Prediction.

Fair machine learning in the legal domain is a relatively new field. Studies such as \citealt{angwin2016machine} identified racial bias in the COMPAS system, a parole risk assessment tool in the US, where black individuals were more likely to be mislabeled as high risk. Another study by \citealt{wang2021equality} found significant fairness gaps across gender in LSTM-based models for legal judgment consistency using a dataset of Chinese criminal cases. Recently, \citealt{chalkidis2022fairlex} developed the FairLex benchmark to facilitate research on bias mitigation algorithms in the legal domain. It includes four datasets from different jurisdictions and languages, covering various sensitive attributes. While previous works focused on group fairness and quantifying predictions across demographic subgroups, this study examines a specific variable of lower court from a counterfactual perspective.

\section{Occlusion \& LCI Dataset for SJP}
The SJP dataset \cite{niklaus2021swiss} comprises 85,000 cases from the Federal Supreme Court of Switzerland (FSCS) spanning the years 2000 to 2020, chronologically split into training (2000-14), validation (2015-16) and test (2017-20) splits and are written in three languages: German, French, and Italian. However, it is important to note that the dataset is not evenly distributed among these languages with Italian having a much smaller number of documents (4K) compared to German (50K) and French (31k). Additionally, this representation disparity is also evident across various legal areas and regions. For more detailed dataset statistics, please refer to the work of \citealt{niklaus2021swiss}.

\subsection{Rationale and Lower Court Annotation}
We sample a total of 108 cases from both the validation and test sets (2015-20). These cases were equally distributed across the three languages. Within each year of the validation and test sets, we sampled six cases per language, resulting in two cases per legal area. Specifically, each legal area in every year contained one case with the judgment "approved" and one with the judgment "dismissed." It is worth noting that our annotation dataset is balanced in terms of final outcomes and languages, in contrast to the SJP dataset, which contains a majority of dismissed cases (> 3/4). The annotations were conducted by a team of three legal experts, consisting of two law students pursuing their master's degrees and one lawyer, over a period of five months. Two legal experts are native German speakers with intermediate knowledge in French and basic Italian skills. The third expert is a native speaker in German and Italian and fluent in French. The annotation was facilitated using the Prodigy tool \footnote{\url{https://prodi.gy}}.

The annotation task was to highlight sentences or sub-sentences in the facts section of the judgment that "support" or "oppose" the final outcome of the case. We have chosen sub-sentences as the atomic unit for annotation after consulting with legal experts who expressed that a sentence can contain two sub-sentences opposing each other and hence should be annotated with different labels. The annotators had been given access to the entire case to make their annotation instead of just the facts section, which is the actual input for the models dealing with judgment prediction task. These decisions have been taken to address two points:
(i)~Experts opined that sentences/sub-sentences may have opposing labels depending on how the court interpreted those facts in its reasoning; hence providing them the entire case would greatly assist them in arriving at explanations leading to higher inter-annotator agreement (ii)~Having prior knowledge about a specific case allows an expert to be familiar with its specific legal and factual details, as well as the court's opinions on the matter. As a result, varying levels of prior familiarity with a case can lead to different interpretations and perspectives in understanding it. Hence providing the entire case levels the playing field and eliminates the possibility that some cases are known to only some experts before, possibly leading to different annotations.

The experts are instructed to read through the facts, the considerations, the ruling, and any other needed legal document (such as relevant legislation, analyses or case law) to understand the court case and then annotate the rationale. 
Unlike the previous works involving explainability annotations in LJP \cite{chalkidis2021paragraph,santosh2022deconfounding,malik2021ildc} which only collect rationales that help to arrive at the final outcome of the case,  we introduce and collect rationales at fine-grained level termed as "Supports Judgement" and  "Opposes Judgment" which holds significance especially in the task of judgment prediction due to the inherent nature of legal text of often operating within the realm of gray areas rather than clear-cut black-and-white distinctions. Legal cases involve complex issues, conflicting facts leading to alternative legal reasoning, dissenting opinions, alternative interpretations of the law and can serve as potential grounds for challenging the ruling and can serve as a reference point for legal arguments or considerations. Thus, including the fine granularity of labels provides an opportunity to assess more  nuanced understanding of the case by the models, acknowledging that legal decisions are not always unanimous and different perspectives may exist within the legal community.  

Additionally, we request annotators to label neutral sentences. This is not a label per se, but covers sentences not assigned other labels, as this assists in implementing the occlusion method to partition the facts section into more coherent sentences with minimal effort, as segmenting legal text is a complex task in itself \cite{read2012sentence,savelka2017sentence,brugger_multilegalsbd_2023}. 

In addition to sentences and sub-sentences indicating towards outcome explanations, we also ask annotators to label the lower court mentions in the fact section as indicated in the rubrum (header including identifiers, and listing judges, lawyers and involved parties) of the ruling. 

The annotation task was conducted in two cycles to ensure high quality. The initial cycle involved pilot annotations, highlighting uncertainties regarding guidelines. As a result, we refined the guidelines by providing more precise instructions to address these concerns\footnote{Our detailed annotation guidelines and discussions are available \hyperlink{https://github.com/santoshTYSS/ExplainabilitySJP_AnnotationGuidelines}{here}.}. Subsequently, a discussion among the legal experts was held to resolve any conflicts and consolidate the annotations in the most effective manner, thereby ensuring the high quality of the annotations.

\begin{figure}
    \centering
    \includegraphics[width = 0.45\textwidth]{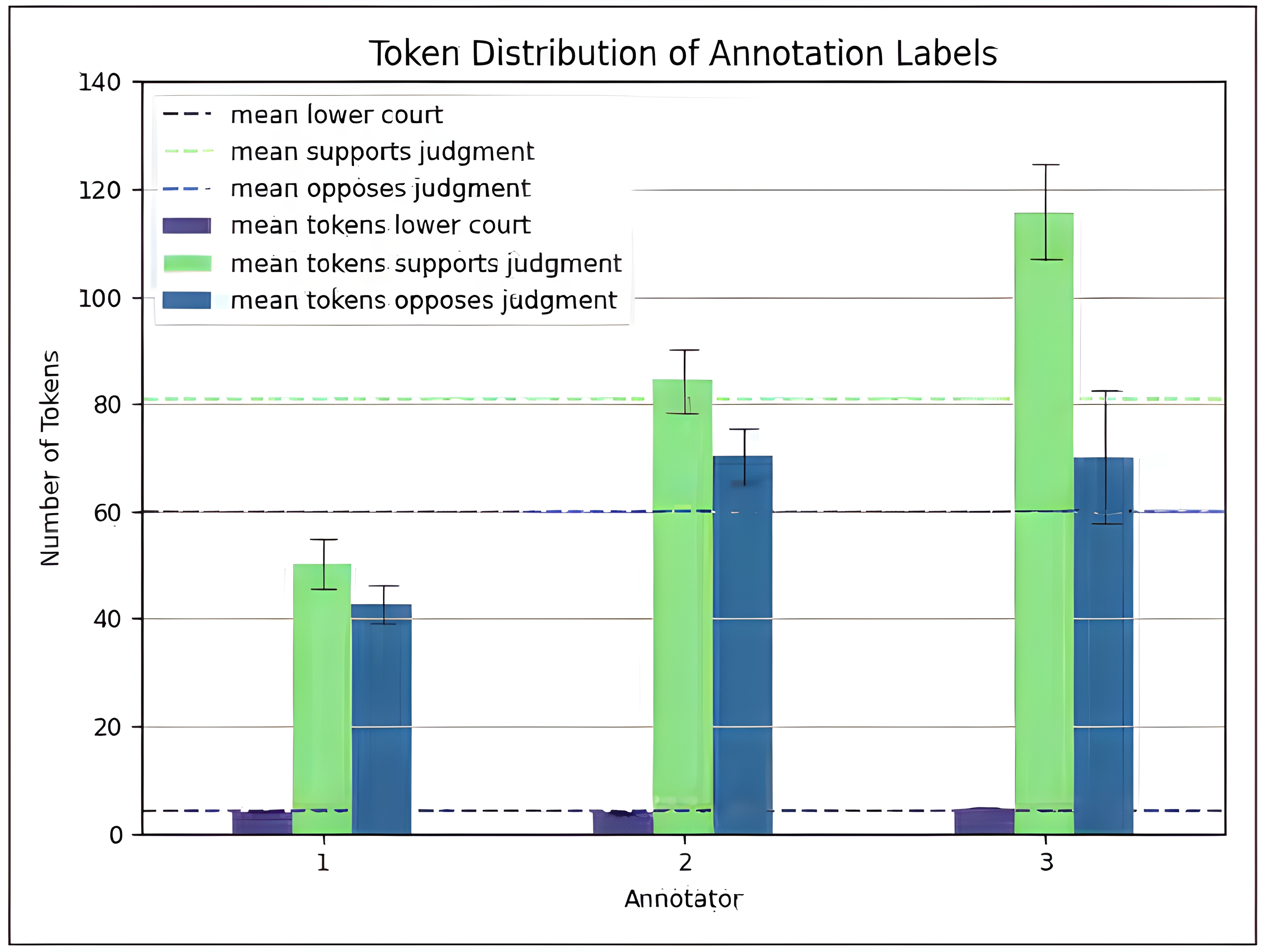}
    \caption{Mean number of tokens annotated per label per annotator in German subset}
    \label{fig:mean-annotation}
\end{figure}

\subsection{Inter Annotator Agreement}
\begin{figure}[t]
    \centering
    \includegraphics[width = 0.45\textwidth]{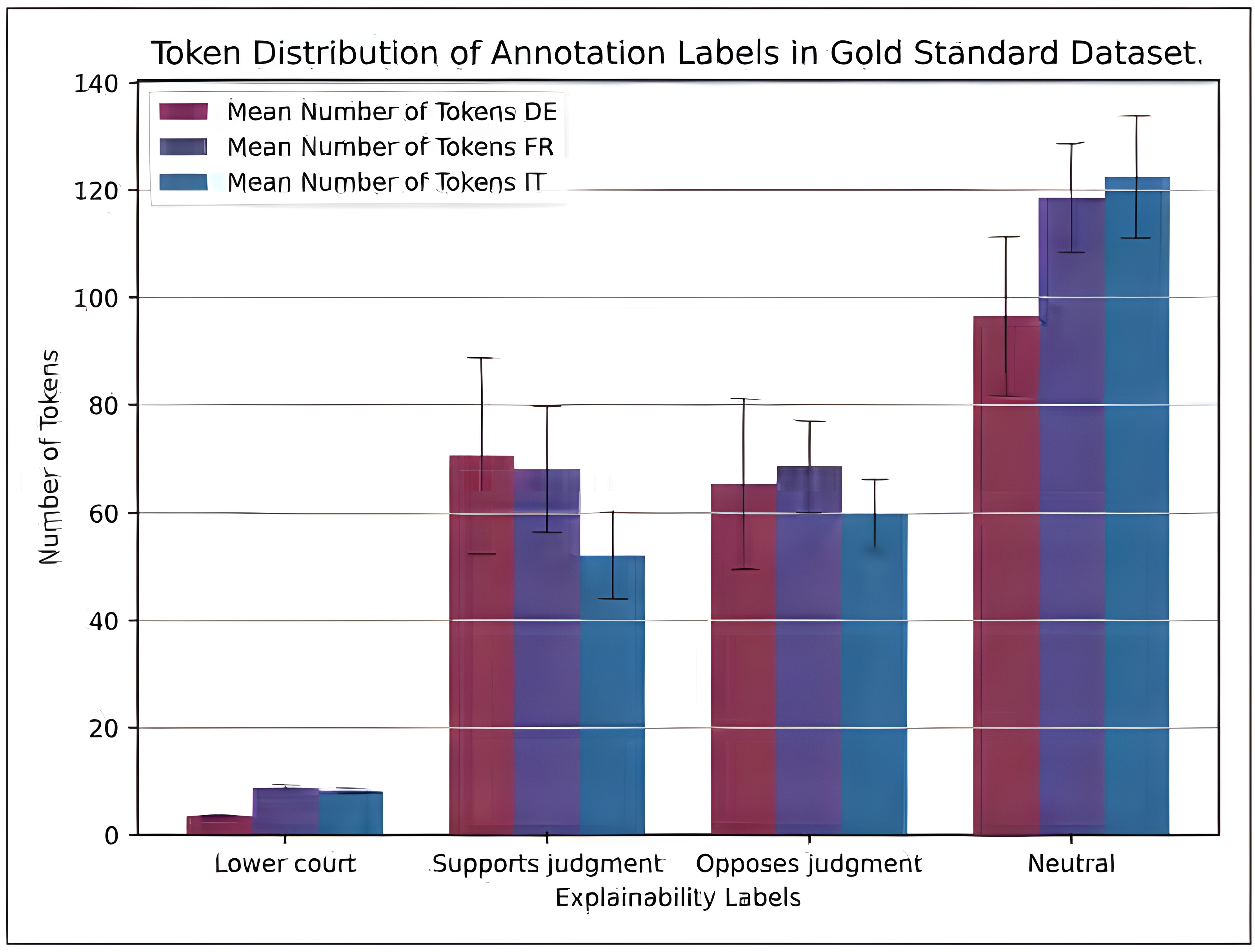}
    \caption{Distribution of the number of tokens per label in the final dataset across each language.}
    \label{fig:tokens_final_lang}
\end{figure}

We obtained annotations from three annotators only for the German subset. Detailed distribution of labeled tokens per annotator can be found in  Fig. \ref{fig:mean-annotation}. Among the three, Annotator 1 annotated the least amount of tokens. Annotator 3 annotated the most comparable to the Annotator 2, especially when using the Supports Judgment label. 
To measure inter-annotator agreement for explanations, we use the machine translation  metrics as suggested by \citealt{malik2021ildc} like ROUGE-1, ROUGE-2, ROUGE-L \cite{lin2004rouge}, BLEU \cite{papineni2002bleu} (unigram and bigram averaging), METEOR \cite{agarwal2007meteor}, Jaccard Similarity, Overlap Maximum, Overlap Minimum and BERTScore \cite{zhang2019bertscore}. We report the inter-annotator agreement scores in the German subset for the first round of annotations of the German dataset in Table \ref{iaa}. These scores are aggregated over all the labels (supports, opposes judgment and lower court). Table \ref{iaa} demonstrates high agreement across all scores, with values ranging from 0.7 to 0.9. The high BERTScore indicates strong similarities in non-lexical matches, while the indication of OVERLAP Minimum suggests that the annotations frequently overlapped as sub-sequences. Notably, Experts 2 and 3 exhibit the highest agreement, which can be attributed to their larger number of annotated tokens compared to Expert 1 as observed in Fig. \ref{fig:mean-annotation}. 
We also notice that the agreement within the categories "Lower Court" and "Supports Judgment" is notably high in comparison to "Opposes Judgment". The experts confirmed that the higher variance in the "Opposes Judgment" label stemmed from the difficulty in identifying these sentences and resolving these conflicts constituted a significant effort in landing with final annotations. Distribution of final number of tokens obtained per label across language is visualised in Fig. \ref{fig:tokens_final_lang}.

\begin{table}[t]
    \centering
    \begin{tabular}{l|c|c|c}
    \toprule
    \textbf{IAA metric} & \textbf{A1-A2} & \textbf{A1-A3} & \textbf{A2-A3} \\
    \midrule
    Rouge-1 & 0.78 & 0.69 & 0.87 \\
    Rouge-2 & 0.74 & 0.64 & 0.85 \\
    Rouge-L & 0.77 & 0.68 & 0.87 \\
    BLEU & 0.75 & 0.69 & 0.85 \\
    METEOR & 0.77 & 0.71 & 0.88 \\
    Jaccard Sim. & 0.73 & 0.64 & 0.82 \\
    Overlap Max. & 0.68 & 0.61 & 0.74 \\ 
    Overlap Min.  & 0.83 & 0.73 & 0.81 \\
    BERTScore  & 0.91  & 0.86 & 0.93 \\
    \bottomrule
    \end{tabular}
    \caption{IAA score between the annotators in the first cycle for German subset}
    \label{iaa}
\end{table}

\subsection{Occlusion and LCI dataset}
To evaluate the explainability of models and enable a fair comparison among them, we derive four distinct occlusion based datasets from the test split\footnote{We exclude the instances from the validation split, which is used for hyperparameter tuning during model training, to derive the occluded test set for explainability.} of above annotated rationales data, consisting of 27, 24 and  23 cases in German, French and Italian respectively. For each occlusion test set, we occlude a different number of sentences (1, 2, 3, and 4)  belonging to same label (Supports/Opposes Judgment/Neutral) per experiment in a case, adding no marker or trace of the occlusion in the fact section to leave it as similar and natural as possible. For every occlusion test instance, we also pair it with a baseline with no text occluded. Thus, we arrive in a total of 28k occluded instances with varying levels of occlusion, across three languages. Using these occluded instances, we analyze the difference in prediction confidence in comparison to the non-occluded baseline. 

For LCI, we derive the counterfactual based test set wherein we use the lower court instances annotated by the annotator and replace the lower court instances with other lower court names in every case resulting in a total of 1127 instances. There are a total of 13, 9, and 16 unique lower court instances in German, French, and Italian respectively. Similar to above, each instance is also paired with a baseline representing the case text with the actual lower court name without any replacement, which we use to analyze the change in prediction confidence. 

Tables \ref{occ_data} and \ref{final_data} provide statistics on the total number of instances in both the Occlusion and LCI test sets along with detailed breakdown of the number of instances in each occlusion set by language. Across languages, the German subset comprises the largest portion of the test set compared to the rest. This is due to annotation of fewer sentences in Italian and French documents as can be noticed in Fig. \ref{fig:tokens_final_lang}.  Among the labels, the `Opposes Judgment' label has fewer instances, due to the lower number of annotated tokens associated with this label.

\begin{table}[]
\begin{tabular}{l|l|c|c|c}
\toprule
\textbf{}      & \textbf{}      & \textbf{DE} & \textbf{FR} & \textbf{IT} \\  \midrule
\multicolumn{2}{l}{\#Documents} & 27          & 24          & 23          \\ \midrule
Set 1          & Opposes        & 55          & 34          & 31          \\
               & Neutral        & 247         & 164         & 195         \\
               & Supports       & 98          & 85          & 50          \\ \midrule
Set 2          & Opposes        & 66          & 22          & 23          \\
               & Neutral        & 1097        & 586         & 827         \\
               & Supports       & 203         & 246         & 69          \\ \midrule
Set 3          & Opposes        & 53          & 7           & 8           \\
               & Neutral        & 3158        & 1260        & 2429        \\
               & Supports       & 356         & 659         & 56          \\ \midrule
Set 4          & Opposes        & 27          & 1           & 1           \\
               & Neutral        & 6622        & 1801        & 5704        \\
               & Supports       & 586         & 1477        & 28        \\  \bottomrule
\end{tabular}
\caption{Split of number of instances per label in each occluded test across three languages.}
\label{occ_data}
\end{table}

\begin{table}[]
\begin{tabular}{l|cccc|c}
\toprule
\textbf{}   & \multicolumn{4}{c|}{\textbf{Occlusion}}                                                                                                 & \textbf{LCI}   \\ \midrule
\textbf{}   & \multicolumn{1}{c|}{\textbf{Opp.}} & \multicolumn{1}{c|}{\textbf{Neu.}} & \multicolumn{1}{c|}{\textbf{Sup.}} & \textbf{Total} & \textbf{Total} \\ \midrule
\textbf{DE} & \multicolumn{1}{c|}{201}              & \multicolumn{1}{c|}{11325}            & \multicolumn{1}{c|}{1243}              & 12769          & 351            \\ 
\textbf{FR} & \multicolumn{1}{c|}{64}               & \multicolumn{1}{c|}{3875}             & \multicolumn{1}{c|}{2467}              & 6406           & 391            \\ 
\textbf{IT} & \multicolumn{1}{c|}{63}               & \multicolumn{1}{c|}{9218}             & \multicolumn{1}{c|}{203}               & 9484           & 312            \\ \bottomrule
\end{tabular}
\caption{Total number of instances per label across three languages for occlusion and LCI test. Opp., Neu., Sup. represent `Opposes Judgement', `Neutral' and `Supports Judgement' respectively.}
\label{final_data}
\end{table}

\section{Experimental Setup}
\subsection{Models}
We assess the following six classes of models, developed on the backbone of hierarchical BERT, developed for the SJP task in previous literature \cite{niklaus2021swiss,niklaus2022empirical}. We follow the same dataset splits provided by \citealt{niklaus2021swiss} for training and validation. Hierarchical BERT is employed because the SJP dataset includes documents with more than 512 tokens. In this approach, the text is split into 4 consecutive blocks of 512 tokens (90\% of cases are less than 2048 tokens) and fed into a shared standard BERT encoder independently. Then the CLS token of each block is passed through a 2-layer transformer encoder to aggregate the information across blocks, followed by max-pooling and a final classification layer.
\newline 

\noindent \textbf{MonoLingual:} This variant uses monolingually pre-trained BERT models i.e German-BERT \cite{chan2019deepset}, CamemBERT \cite{martin2020camembert} and UmBERTo \cite{parisi2020umberto} for German, French and Italian. Each model is fine-tuned and evaluated using that language subset dataset.
\newline 

\noindent \textbf{MultiLingual:} This variant uses the multilingually pre-trained XLM-R model \cite{conneau2019unsupervised} instead of language-specific pre-trained BERT. However, the fine-tuning process is still performed separately for each language, similar to the monolingual approach.
\newline 

\noindent \textbf{Mono/Multi Lingual with Data Augmentation:} We translate the cases in SJP dataset into other languages from the original language using the EasyNMT2 framework, following the approach proposed by \citealt{niklaus2022empirical}. Then these translated instances are then augmented with the original data for a specific language during the fine-tuning process with Mono/Multilingual BERT. This is similar to above experiment in setup, with the main distinction being the additional augmented data. 
\newline 

\noindent \textbf{Joint Training without/with Data Augmentation:} We use a multilingual pre-trained model and fine-tune it across all the three language corpora jointly, which tries to capitalize on the inherited benefit of using larger multilingual corpora during fine-tuning. 
As discussed above, we translate each document from its original language to the rest of other two and train a data augmented version of models jointly with all the  obtained translated data. 
Unlike the previous approaches where separate models were fine-tuned for each language, this method jointly fine-tunes on all languages, resulting in a single final model instead of multiple models for each language.


\subsection{Implementation Details}
\label{impl}
We use the code repositories from prior work \cite{niklaus2021swiss,niklaus2022empirical} to assess the state-of-the-art models on SJP\footnote{\url{https://github.com/JoelNiklaus/SwissJudgementPrediction}}. We employ a learning rate of 1e-5 with early stopping based on macro-F1 on the development set. All models are trained with a batch size of 64 for 10 epochs using AdamW optimizer with mixed precision and gradient accumulation using huggingface library \cite{wolf2020transformers}. We use oversampling to handle class imbalance. We use 4 segments with 512 tokens each in our hierarchical models resulting in a maximum sequence length of 2048. 

\subsection{Metrics}
We report macro-F1 following \citealt{niklaus2021swiss,niklaus2022empirical} for assessing prediction performance. For assessing explainability through occlusion experiments, we calculate the explainability score $S_{exp}$ for every test instance as the difference between the temperature-scaled\footnote{We adopt temperature scaling \cite{guo2017calibration} to calibrate the confidence estimates of the model. 
} confidence 
of the baseline and the occluded instance. (i.e., baseline - occluded).
A negative (positive) $S_{exp}$ score indicates that occluded text is opposing (supporting) its prediction. Then, we assign the label `Opposes Judgement'/`Neutral'/'Supports Judgement' based on the sign of explainability score. 
Finally, we report F1-score for each of the labels across all the occluded instances.

In bias estimation using the LCI method, we calculate an explainability score for each instance. As the explainability scores are sign dependent, we separately compute the Mean of Explainability Scores (MES), for positive and negative values, expressed as a percentage. A positive explainability score indicates that the insertion of the lower court decreases the probability, suggesting that the inserted court has a pro-dismissal influence. Conversely, a negative score indicates an increase in the probability, indicating a pro-approval trend of the inserted lower court. For an ideally unbiased model, the presence of the lower court should not affect the probability of the prediction. Therefore, a value of the mean explainability score closer to 0 is desirable. Additionally, we report the percentage of cases where the insertion of the lower court leads to a flip in the label of the prediction, changing it from 0 to 1 or vice versa. 

\begin{table}[]
\centering
\scalebox{0.9}{
\begin{tabular}{lccc}
\toprule
\textbf{Model}             & \multicolumn{1}{l|}{\textbf{German}} & \multicolumn{1}{l|}{\textbf{French}} & \multicolumn{1}{l|}{\textbf{Italian}} \\ \midrule
MonoLingual       & 69.08                       & 71.78                       & 67.82                        \\ \midrule
MultiLingual      & 67.92                       & 69.24                       & 65.28                        \\ \midrule
MonoLingual + DA  & 70.47                       & 71.24                       & 69.21                        \\ \midrule
MultiLingual + DA & 68.94                       & 71.06                       & 69.86                        \\ \midrule
Joint Training      & 68.74                       & 70.82                       & 70.62                        \\ \midrule
Joint Training + DA & 70.58                       & 71.62                       & 71.18                        \\ \bottomrule
\end{tabular}}
\caption{Prediction Performance on Test set of \citealt{niklaus2021swiss}}
\label{pred}
\end{table}

\begin{table*}[]
\centering
\scalebox{0.85}{
\begin{tabular}{|l|ccc|ccc|ccc|}
\toprule
Model             & \multicolumn{3}{c|}{German}                                                                          & \multicolumn{3}{c|}{French}                                                                          & \multicolumn{3}{c|}{Italian}                                                                        \\ 
                  & \multicolumn{1}{c|}{Opposes}          & \multicolumn{1}{c|}{Neutral} & \multicolumn{1}{c|}{Supports} & \multicolumn{1}{c|}{Opposes}          & \multicolumn{1}{c|}{Neutral} & \multicolumn{1}{c|}{Supports} & \multicolumn{1}{c|}{Opposes}         & \multicolumn{1}{c|}{Neutral} & \multicolumn{1}{c|}{Supports} \\ \midrule
MonoLingual       & \multicolumn{1}{c|}{3.02 } & \multicolumn{1}{c|}{16.78}        &                         15.10      & \multicolumn{1}{c|}{1.95}   & \multicolumn{1}{c|}{3.68}        &   40.24                          & \multicolumn{1}{c|}{0.49 }  & \multicolumn{1}{c|}{3.68 }        & 11.24                       \\ \midrule
MultiLingual      & \multicolumn{1}{c|}{2.04 } & \multicolumn{1}{c|}{11.90 }        &  17.46                           & \multicolumn{1}{c|}{1.77}  & \multicolumn{1}{c|}{3.62}        &   42.77                            & \multicolumn{1}{c|}{0.85 } & \multicolumn{1}{c|}{5.72 }        &  13.48                            \\ \midrule
MonoLingual + DA  & \multicolumn{1}{c|}{3.21 } & \multicolumn{1}{c|}{16.26 }        &   18.08                         & \multicolumn{1}{c|}{1.78}  & \multicolumn{1}{c|}{5.98}        &     43.12                         & \multicolumn{1}{c|}{0.98 } & \multicolumn{1}{c|}{4.39}       &   14.99                           \\ \midrule
MultiLingual + DA & \multicolumn{1}{c|}{3.64 } & \multicolumn{1}{c|}{19.06}        &   20.83                           & \multicolumn{1}{c|}{1.43}  & \multicolumn{1}{c|}{4.63}        &      45.77                      & \multicolumn{1}{c|}{0.83 } & \multicolumn{1}{c|}{4.84 }        &   15.36                           \\ \midrule
Joint Training      & \multicolumn{1}{c|}{2.62 } & \multicolumn{1}{c|}{15.72}        & 26.97                           & \multicolumn{1}{c|}{1.67 }   & \multicolumn{1}{c|}{4.19 }        &   48.51                          & \multicolumn{1}{c|}{0.54} & \multicolumn{1}{c|}{5.37 }        &    18.82                        \\ \midrule
Joint Training + DA & \multicolumn{1}{c|}{3.75} & \multicolumn{1}{c|}{14.54}        & 21.95                             & \multicolumn{1}{c|}{1.93} & \multicolumn{1}{c|}{5.73}        &   45.73                            & \multicolumn{1}{c|}{0.63 } & \multicolumn{1}{c|}{4.82}        &  19.68                          \\ \bottomrule
\end{tabular}}
\caption{Analysis of Explainability using Occlusion -  F1-scores across all instances from all four test sets for each label in every language. Higher the scores, better the explainability. }
\label{occ}
\end{table*}

\begin{table*}[]
\centering
\scalebox{0.73}{
\begin{tabular}{|l|llll|llll|llll|}
\toprule
\multicolumn{1}{|c|}{Model} & \multicolumn{4}{c|}{German}         & \multicolumn{4}{c|}{French}&\multicolumn{4}{c|}{Italian}        \\ \midrule
\multicolumn{1}{|c|}{}      & 
\multicolumn{1}{c|}{\begin{tabular}[c]{@{}c@{}}+ \\ MES\end{tabular}} & \multicolumn{1}{c|}{\begin{tabular}[c]{@{}c@{}}-\\ MES\end{tabular}} & \multicolumn{1}{c|}{\begin{tabular}[c]{@{}c@{}}Flip \\ 1 $\rightarrow$ 0\end{tabular}} & \multicolumn{1}{c|}{\begin{tabular}[c]{@{}c@{}}Flip \\ 0 $\rightarrow$ 1\end{tabular}} & \multicolumn{1}{c|}{\begin{tabular}[c]{@{}c@{}}+ \\ MES\end{tabular}} & \multicolumn{1}{c|}{\begin{tabular}[c]{@{}c@{}}-\\ MES\end{tabular}} & \multicolumn{1}{c|}{\begin{tabular}[c]{@{}c@{}}Flip \\ 1 $\rightarrow$ 0\end{tabular}} & \multicolumn{1}{c|}{\begin{tabular}[c]{@{}c@{}}Flip \\0 $\rightarrow$ 1\end{tabular}} & \multicolumn{1}{c|}{\begin{tabular}[c]{@{}c@{}}+ \\ MES\end{tabular}} & \multicolumn{1}{c|}{\begin{tabular}[c]{@{}c@{}}-\\ MES\end{tabular}} & \multicolumn{1}{c|}{\begin{tabular}[c]{@{}c@{}}Flip \\ 1 	$\rightarrow$
0\end{tabular}} & \multicolumn{1}{c|}{\begin{tabular}[c]{@{}c@{}}Flip \\ 0 $\rightarrow$ 1\end{tabular}} \\ \midrule
MonoLingual & \multicolumn{1}{l|}{$3.48_{5.12}$}           & \multicolumn{1}{l|}{$-2.3_{2.82}$}                                            & \multicolumn{1}{l|}{2.28}                                                              & 0.43                                                                                   & \multicolumn{1}{l|}{$2.56_{2.65}$}               & \multicolumn{1}{l|}{$-2.2_{3.79}$}                                            & \multicolumn{1}{l|}{3.84}                                                              & 1.02                                                                                   & \multicolumn{1}{l|}{$1.64_{3.46}$}              & \multicolumn{1}{l|}{$-2.22_{7.76}$}                                           & \multicolumn{1}{l|}{0.12}                                                              & 1.20                                                                                   \\ \midrule
MultiLingual                & \multicolumn{1}{l|}{$3.39_{5.43}$}              & \multicolumn{1}{l|}{$-2.77_{3.64}$}                                           & \multicolumn{1}{l|}{1.71}                                                              & 0.32                                                                                   & \multicolumn{1}{l|}{$4.01_{4.62}$}              & \multicolumn{1}{l|}{$-3.06_{3.22}$}                                                & \multicolumn{1}{l|}{0.51}                                                              & 0.12                                                                                   & \multicolumn{1}{l|}{$1.72_{2.18}$}               & \multicolumn{1}{l|}{$-1.78_{2.52}$}                                           & \multicolumn{1}{l|}{1.05}                                                              & 2.88                                                                                   \\ \midrule
MonoLingual + DA            & \multicolumn{1}{l|}{$3.09_{5.15}$}              & \multicolumn{1}{l|}{$-2.77_{5.42}$}                                           & \multicolumn{1}{l|}{2.56}                                                              & 0.22                                                                                   & \multicolumn{1}{l|}{$4.12_{6.73}$}              & \multicolumn{1}{l|}{$-1.83_{2.34}$}                                                & \multicolumn{1}{l|}{1.24}                                                              & 0.82                                                                                   & \multicolumn{1}{l|}{$1.25_{2.23}$}              & \multicolumn{1}{l|}{$-1.29_{1.97}$}                                           & \multicolumn{1}{l|}{0.32}                                                              & 2.19                                                                                   \\ \midrule
MultiLingual + DA           & \multicolumn{1}{l|}{$5.32_{8.27}$}                & \multicolumn{1}{l|}{$-3.35_{5.24}$}                                           & \multicolumn{1}{l|}{4.56}                                                              & 2.56                                                                                   & \multicolumn{1}{l|}{$4.08_{6.05}$}              & \multicolumn{1}{l|}{$-6.48_{9.64}$}                                                & \multicolumn{1}{l|}{1.53}                                                              & 3.07                                                                                   & \multicolumn{1}{l|}{$2.88_{3.51}$}               & \multicolumn{1}{l|}{$-6.12_{8.08}$}                                           & \multicolumn{1}{l|}{2.56}                                                              & 3.85                                                                                   \\ \midrule
Joint Training                & \multicolumn{1}{l|}{$3.32_{6.18}$}              & \multicolumn{1}{l|}{$-1.86_{2.89}$}                                           & \multicolumn{1}{l|}{3.13}                                                              & 1.99                                                                                   & \multicolumn{1}{l|}{$4.08_{4.37}$}              & \multicolumn{1}{l|}{$-2.71_{3.69}$}                                                & \multicolumn{1}{l|}{0.51}                                                              & 2.56                                                                                   & \multicolumn{1}{l|}{$6.13_{6.79}$}              & \multicolumn{1}{l|}{$-2.66_{2.12}$}                                           & \multicolumn{1}{l|}{4.92}                                                              & 4.24                                                                                   \\ \midrule
Joint Training + DA           & \multicolumn{1}{l|}{$3.23_{4.46}$}              & \multicolumn{1}{l|}{$-1.84_{2.45}$}                                           & \multicolumn{1}{l|}{2.85}                                                              & 1.99                                                                                   & \multicolumn{1}{l|}{$3.04_{3.96}$}               & \multicolumn{1}{l|}{$-4.04_{4.31}$}                                                & \multicolumn{1}{l|}{3.07}                                                              & 2.32                                                                                   & \multicolumn{1}{l|}{$6.14_{7.94}$}              & \multicolumn{1}{l|}{$-3.21_{3.11}$}                                           & \multicolumn{1}{l|}{4.09}                                                              & 4.83                                                                                   \\ \bottomrule
\end{tabular}}
\caption{Analysis of Lower Court Bias using LCI - Results of Positive and Negative MES Scores, and Label Flips across the three languages. Labels 0 and 1 indicates dismissal and approval respectively. Lower scores indicate a less biased model. Subscript indicates the standard deviation values.}
\label{lci}
\end{table*}

\subsection{Analysis using Occlusion Test}
We present the results of prediction performance and the explainability analysis using occlusion in Tables \ref{pred} and \ref{occ}, respectively. Analyzing Table \ref{occ}, we observe that the model achieves higher accuracy in classifying instances with Supports Judgment compared to those with Neutral or Opposes Judgment. This could be attributed to the fact that the Opposes Judgment category is underrepresented in the occlusion dataset (due to fewer annotated tokens with this label) and the challenging task to classify Neutral instances. Among the three languages, French exhibits the highest score for the Supports Judgment category, but it also shows lower scores for the other classes.

Despite the MultiLingual model displaying a decrease in predictive performance, it shows some improvement in occlusion performance, particularly for the Supports Judgments class, across all languages. A similar trend is observed in the Joint training model, which consistently demonstrates a significant increase in explainability scores across languages for most classes.

While the inclusion of the DA component in both MonoLingual and MultiLingual models resulted in improved explainability scores for most labels compared to their counterparts, its addition to the Joint training model leads to mixed results. Surprisingly, the addition of the DA component to the Joint training model consistently increases prediction performance but does not maintain consistency in explainability performance. This finding emphasizes the importance of evaluating explainability to develop transparent systems that can make accurate predictions for the right reasons.  

Overall, the lower scores across the board indicate the flawed inference about factors predictive for the outcome. Despite the impressive performances of state-of-the-art models on standard LJP prediction performance, there is still much progress to be made to make those models align as closely as possible with the rationales deemed relevant by legal experts. To create practical value for the legal field, the field of LJP should aim for a productive fusion of expert knowledge and data-driven insights, rather than data-driven correlation based learning.  

\subsection{Analysis using LCI Test}
From Table \ref{lci}, we can observe that the modification of the lower court has a considerable influence on the overall prediction confidence, as indicated by the changes in confidence scores up to 5\% in both directions across all languages, despite the lower court name on average spanning around seven tokens in documents of an average length of 350 tokens. However, these small changes in confidence scores did result in label flips.

Overall, no consensus exists on which model setting has yielded lower MES scores in both directions consistently across all the languages. MultiLingual model's prediction performance decreased compared to the MonoLingual model across all three languages and its bias scores increased significantly across all languages, barring -MES for Italian and +MES for German. While the inclusion of the DA (Data Augmentation) component resulted in improved prediction performance compared to the non-DA variants in both MonoLingual and MultiLingual settings, the Multilingual + DA model exhibited a notable increase in bias. This suggests that the model's reliance on lower court names is more pronounced in the presence of the DA component compared to its non-DA variant.

Joint training models, which aim to generalize across languages, demonstrate improved prediction performance, particularly for Italian, which is underrepresented in the training set. However, this improvement comes at the cost of higher MES scores, indicating potential overfitting to court-specific correlations rather than capturing the actual reasoning behind the predictions.
Interestingly, despite training a single model using data from all three languages, the Italian data shows a significant divergence in MES scores compared to German and French. This highlights that representational bias across languages seems to be a crucial part.  While adding DA to Joint training models significantly improves prediction performance, bias scores lack a clear pattern.

Across all the models in the German setting, we can witness an overall pro-dismissal trend (greater + MES scores compared to - MES) echoing with more number of label flips from approval to dismissal. While French notices an overall pro-dismissal trend,  Italian shows pro-approval trend, barring the cross-lingual models.  These observed biases regarding lower court underscore the need for continuous bias evaluation and mitigation in LJP models. 

\section{Conclusion}
In this work, we present the rationale dataset curated at fine-grained level of both `supporting' and 'opposing' factors for Swiss Judgment Prediction (SJP), the only available multilingual LJP dataset. We employ a perturbation-based occlusion approach to assess various state-of-the-art models developed for SJP and also release four distinct occlusion test sets, occluding a different number of sentences in each of the sets. Our lower explainability scores suggest that the current models do not align well with the legal experts which can lead to sub-optimal litigation strategies due to flawed inference about factors responsible for the outcome. Furthermore, we assess the bias of the lower court information in  the final predictions generated by the models using LCI test and notice that models learn spurious correlations about court-outcome in the data. 
In the future, we would explore deconfounding strategy \cite{santosh2022deconfounding} to improve the alignment between what models and experts deem relevant. One can explore different group robust algorithms such as adversarial removal, IRM, Group DRO and V-REx, as an effective bias-mitigation strategy \cite{chalkidis2022fairlex} and investigate its impact on explainability. We hope our data resource  will be useful to the research community working on Legal Judgement Prediction.


\section{Limitations} 
In this study, our approach to obtaining rationales involved a consolidation process wherein we aimed to achieve a final set of high-quality annotations through discussions with legal experts. However, it is important to acknowledge that the assumption of a single ground truth may overlook the presence of genuine human variation, which can arise due to factors such as disagreement, subjectivity in annotation, or the existence of multiple plausible answers \cite{xu2023dissonance}. Particularly in the field of law, where complexity and interpretation are inherent, it is well-recognized that lawyers may have differing legal assessments of case facts and how they contribute to the eventual outcome. Instead of attempting to resolve variations in expert labels, it is essential to acknowledge and embrace the inherent variation in human annotations. Moving forward, it is crucial to develop methods that can comprehensively capture and account for variation from data to evaluation, enabling a more comprehensive treatment of this variability in future research.

In the evaluation of our occlusion-based explainability setup, we utilized the F1-score, which focuses solely on the final label obtained from the change in confidence score between the baseline and occluded instances. However, it is important to emphasize the need for a metric that takes into account the magnitude of the difference in confidence scores during aggregation, in order to present a more comprehensive and holistic assessment.

\section{Ethics Statement} 
The dataset used in this work comes from prior work by \citealt{niklaus2021swiss} and these are publicly available on the \url{https://entscheidsuche.ch} platform and the names of the parties have been redacted by the court to ensure anonymity. 

This work does not endorse or advocate for practical use of such systems. Instead our aim in this work is to rather empirically demonstrate that these systems are far from practical use due to their flawed inference about factors leading to outcome prediction. The scope of this work is to study LJP from an explainability standpoint and to showcase the discrepancy between the prediction performance and explainability performance and emphasize the need to build technology that can help practitioners with reliable insights. Our dataset and findings associated with this work will contribute to advancing the field of explainable legal judgement prediction and provide valuable insights for developing more reliable and unbiased models in the future. 

Furthermore, we would like to draw attention to the work by \citealt{tsarapatsanis2021ethical} which discusses various normative factors related to ethics in the context of legal natural language processing. These discussions are crucial for fostering ethical thinking within the legal NLP community and ensuring the responsible development of systems that can assist lawyers, judges, and the general public.

\section*{Acknowledgements}
Thanks to Ilias Chalkidis for providing feedback on initial ideas and experiments and to Lynn Grau, Thomas Lüthi \& Angela Stefanelli for providing expert legal annotations.

\section{Bibliographical References}\label{sec:reference}

\bibliographystyle{lrec-coling2024-natbib}
\bibliography{lrec-coling2024}

\begin{thebibliography}{74}
\expandafter\ifx\csname natexlab\endcsname\relax\def\natexlab#1{#1}\fi

\bibitem[{Agarwal and Lavie(2007)}]{agarwal2007meteor}
Abhaya Agarwal and Alon Lavie. 2007.
\newblock Meteor: An automatic metric for mt evaluation with high levels of
  correlation with human judgments.
\newblock \emph{Proceedings of WMT-08}.

\bibitem[{Aletras et~al.(2016)Aletras, Tsarapatsanis, Preo{\c{t}}iuc-Pietro,
  and Lampos}]{aletras2016predicting}
Nikolaos Aletras, Dimitrios Tsarapatsanis, Daniel Preo{\c{t}}iuc-Pietro, and
  Vasileios Lampos. 2016.
\newblock Predicting judicial decisions of the european court of human rights:
  A natural language processing perspective.
\newblock \emph{PeerJ Computer Science}, 2:e93.

\bibitem[{Aleven and Ashley(1997)}]{aleven1997teaching}
Vincent Aleven and Kevin~D Ashley. 1997.
\newblock Teaching case-based argumentation through a model and examples:
  Empirical evaluation of an intelligent learning environment.
\newblock In \emph{Artificial intelligence in education}, volume~39, pages
  87--94. Citeseer.

\bibitem[{Angwin et~al.(2016)Angwin, Larson, Mattu, and
  Kirchner}]{angwin2016machine}
Julia Angwin, Jeff Larson, Surya Mattu, and Lauren Kirchner. 2016.
\newblock Machine bias: There’s software used across the country to predict
  future criminals.
\newblock \emph{And it’s biased against blacks. ProPublica}, 23:77--91.

\bibitem[{Bertalan and Ruiz(2020)}]{bertalan2020predicting}
Vithor Gomes~Ferreira Bertalan and Evandro Eduardo~Seron Ruiz. 2020.
\newblock Predicting judicial outcomes in the brazilian legal system using
  textual features.
\newblock In \emph{DHandNLP@ PROPOR}, pages 22--32.

\bibitem[{Brugger et~al.(2023)Brugger, Stürmer, and
  Niklaus}]{brugger_multilegalsbd_2023}
Tobias Brugger, Matthias Stürmer, and Joel Niklaus. 2023.
\newblock \href {https://doi.org/10.48550/arXiv.2305.01211} {{MultiLegalSBD}:
  {A} {Multilingual} {Legal} {Sentence} {Boundary} {Detection} {Dataset}}.
\newblock ArXiv:2305.01211 [cs].

\bibitem[{Br{\"u}ninghaus and Ashley(2003)}]{bruninghaus2003combining}
Stefanie Br{\"u}ninghaus and Kevin~D Ashley. 2003.
\newblock Combining case-based and model-based reasoning for predicting the
  outcome of legal cases.
\newblock In \emph{Case-Based Reasoning Research and Development: 5th
  International Conference on Case-Based Reasoning, ICCBR 2003 Trondheim,
  Norway, June 23--26, 2003 Proceedings 5}, pages 65--79. Springer.

\bibitem[{Br{\"u}ninghaus and Ashley(2005)}]{bruninghaus2005generating}
Stefanie Br{\"u}ninghaus and Kevin~D Ashley. 2005.
\newblock Generating legal arguments and predictions from case texts.
\newblock In \emph{Proceedings of ICAIL 2005}, pages 65--74.

\bibitem[{Chalkidis et~al.(2019)Chalkidis, Androutsopoulos, and
  Aletras}]{chalkidis2019neural}
Ilias Chalkidis, Ion Androutsopoulos, and Nikolaos Aletras. 2019.
\newblock Neural legal judgment prediction in english.
\newblock In \emph{Proceedings of ACL 2019}, pages 4317--4323.

\bibitem[{Chalkidis et~al.(2022{\natexlab{a}})Chalkidis, Dai, Fergadiotis,
  Malakasiotis, and Elliott}]{chalkidis2022exploration}
Ilias Chalkidis, Xiang Dai, Manos Fergadiotis, Prodromos Malakasiotis, and
  Desmond Elliott. 2022{\natexlab{a}}.
\newblock An exploration of hierarchical attention transformers for efficient
  long document classification.
\newblock \emph{arXiv preprint arXiv:2210.05529}.

\bibitem[{Chalkidis et~al.(2021)Chalkidis, Fergadiotis, Tsarapatsanis, Aletras,
  Androutsopoulos, and Malakasiotis}]{chalkidis2021paragraph}
Ilias Chalkidis, Manos Fergadiotis, Dimitrios Tsarapatsanis, Nikolaos Aletras,
  Ion Androutsopoulos, and Prodromos Malakasiotis. 2021.
\newblock Paragraph-level rationale extraction through regularization: A case
  study on european court of human rights cases.
\newblock In \emph{Proceedings of the NAACL-HLT 2021}, pages 226--241.

\bibitem[{Chalkidis et~al.(2022{\natexlab{b}})Chalkidis, Jana, Hartung,
  Bommarito, Androutsopoulos, Katz, and Aletras}]{chalkidis2022lexglue}
Ilias Chalkidis, Abhik Jana, Dirk Hartung, Michael Bommarito, Ion
  Androutsopoulos, Daniel Katz, and Nikolaos Aletras. 2022{\natexlab{b}}.
\newblock Lexglue: A benchmark dataset for legal language understanding in
  english.
\newblock In \emph{Proceedings of ACL 2022}, pages 4310--4330.

\bibitem[{Chalkidis et~al.(2022{\natexlab{c}})Chalkidis, Pasini, Zhang, Tomada,
  Schwemer, and S{\o}gaard}]{chalkidis2022fairlex}
Ilias Chalkidis, Tommaso Pasini, Sheng Zhang, Letizia Tomada, Sebastian
  Schwemer, and Anders S{\o}gaard. 2022{\natexlab{c}}.
\newblock Fairlex: A multilingual benchmark for evaluating fairness in legal
  text processing.
\newblock In \emph{Proceedings of the 60th Annual Meeting of the Association
  for Computational Linguistics (Volume 1: Long Papers)}, pages 4389--4406.

\bibitem[{Chan et~al.(2019)Chan, M{\"o}ller, Pietsch, Soni, and
  Yeung}]{chan2019deepset}
Branden Chan, Timo M{\"o}ller, Malte Pietsch, Tanay Soni, and Chin~Man Yeung.
  2019.
\newblock Deepset-open sourcing german bert.

\bibitem[{Condevaux and Harispe(2022)}]{condevaux2022lsg}
Charles Condevaux and S{\'e}bastien Harispe. 2022.
\newblock Lsg attention: Extrapolation of pretrained transformers to long
  sequences.
\newblock \emph{arXiv preprint arXiv:2210.15497}.

\bibitem[{Conneau et~al.(2019)Conneau, Khandelwal, Goyal, Chaudhary, Wenzek,
  Guzm{\'a}n, Grave, Ott, Zettlemoyer, and Stoyanov}]{conneau2019unsupervised}
Alexis Conneau, Kartikay Khandelwal, Naman Goyal, Vishrav Chaudhary, Guillaume
  Wenzek, Francisco Guzm{\'a}n, Edouard Grave, Myle Ott, Luke Zettlemoyer, and
  Veselin Stoyanov. 2019.
\newblock Unsupervised cross-lingual representation learning at scale.
\newblock \emph{arXiv preprint arXiv:1911.02116}.

\bibitem[{Danilevsky et~al.(2020)Danilevsky, Qian, Aharonov, Katsis, Kawas, and
  Sen}]{danilevsky2020survey}
Marina Danilevsky, Kun Qian, Ranit Aharonov, Yannis Katsis, Ban Kawas, and
  Prithviraj Sen. 2020.
\newblock A survey of the state of explainable ai for natural language
  processing.
\newblock In \emph{Proceedings of the 1st Conference of the Asia-Pacific
  Chapter of the Association for Computational Linguistics and the 10th
  International Joint Conference on Natural Language Processing}, pages
  447--459.

\bibitem[{Grabmair(2017)}]{grabmair2017predicting}
Matthias Grabmair. 2017.
\newblock Predicting trade secret case outcomes using argument schemes and
  learned quantitative value effect tradeoffs.
\newblock In \emph{Proceedings of the 16th edition of the International
  Conference on Articial Intelligence and Law}, pages 89--98.

\bibitem[{Guo et~al.(2017)Guo, Pleiss, Sun, and
  Weinberger}]{guo2017calibration}
Chuan Guo, Geoff Pleiss, Yu~Sun, and Kilian~Q Weinberger. 2017.
\newblock On calibration of modern neural networks.
\newblock In \emph{International conference on machine learning}, pages
  1321--1330. PMLR.

\bibitem[{Hardt et~al.(2016)Hardt, Price, and Srebro}]{hardt2016equality}
Moritz Hardt, Eric Price, and Nati Srebro. 2016.
\newblock Equality of opportunity in supervised learning.
\newblock \emph{Advances in neural information processing systems}, 29.

\bibitem[{Hua et~al.(2022)Hua, Zhang, Chen, Li, and
  Weber}]{hua2022legalrelectra}
Wenyue Hua, Yuchen Zhang, Zhe Chen, Josie Li, and Melanie Weber. 2022.
\newblock Legalrelectra: Mixed-domain language modeling for long-range legal
  text comprehension.
\newblock \emph{arXiv preprint arXiv:2212.08204}.

\bibitem[{Katz et~al.(2017)Katz, Bommarito, and Blackman}]{katz2017general}
Daniel~Martin Katz, Michael~J Bommarito, and Josh Blackman. 2017.
\newblock A general approach for predicting the behavior of the supreme court
  of the united states.
\newblock \emph{PloS one}, 12(4):e0174698.

\bibitem[{Kaufman et~al.(2019)Kaufman, Kraft, and Sen}]{kaufman2019improving}
Aaron~Russell Kaufman, Peter Kraft, and Maya Sen. 2019.
\newblock Improving supreme court forecasting using boosted decision trees.
\newblock \emph{Political Analysis}, 27(3):381--387.

\bibitem[{Kaur and Bozic(2019)}]{kaur2019convolutional}
Arshdeep Kaur and Bojan Bozic. 2019.
\newblock Convolutional neural network-based automatic prediction of judgments
  of the european court of human rights.
\newblock In \emph{AICS}, pages 458--469.

\bibitem[{Kowsrihawat et~al.(2018)Kowsrihawat, Vateekul, and
  Boonkwan}]{kowsrihawat2018predicting}
Kankawin Kowsrihawat, Peerapon Vateekul, and Prachya Boonkwan. 2018.
\newblock Predicting judicial decisions of criminal cases from thai supreme
  court using bi-directional gru with attention mechanism.
\newblock In \emph{2018 5th Asian Conference on Defense Technology (ACDT)},
  pages 50--55. IEEE.

\bibitem[{Lage-Freitas et~al.(2022)Lage-Freitas, Allende-Cid, Santana, and
  Oliveira-Lage}]{lage2022predicting}
Andr{\'e} Lage-Freitas, H{\'e}ctor Allende-Cid, Orivaldo Santana, and
  L{\'\i}via Oliveira-Lage. 2022.
\newblock Predicting brazilian court decisions.
\newblock \emph{PeerJ Computer Science}, 8:e904.

\bibitem[{Li et~al.(2016)Li, Monroe, and Jurafsky}]{li2016understanding}
Jiwei Li, Will Monroe, and Dan Jurafsky. 2016.
\newblock Understanding neural networks through representation erasure.
\newblock \emph{arXiv preprint arXiv:1612.08220}.

\bibitem[{Lin(2004)}]{lin2004rouge}
Chin-Yew Lin. 2004.
\newblock Rouge: A package for automatic evaluation of summaries.
\newblock In \emph{Text summarization branches out}, pages 74--81.

\bibitem[{Liu and Chen(2017)}]{liu2017predictive}
Zhenyu Liu and Huanhuan Chen. 2017.
\newblock A predictive performance comparison of machine learning models for
  judicial cases.
\newblock In \emph{2017 IEEE Symposium series on computational intelligence
  (SSCI)}, pages 1--6. IEEE.

\bibitem[{Lundberg and Lee(2017)}]{lundberg2017unified}
Scott~M Lundberg and Su-In Lee. 2017.
\newblock A unified approach to interpreting model predictions.
\newblock \emph{Advances in neural information processing systems}, 30.

\bibitem[{Luo et~al.(2017)Luo, Feng, Xu, Zhang, and Zhao}]{luo2017learning}
Bingfeng Luo, Yansong Feng, Jianbo Xu, Xiang Zhang, and Dongyan Zhao. 2017.
\newblock Learning to predict charges for criminal cases with legal basis.
\newblock In \emph{Proceedings of EMNLP 2017}, pages 2727--2736.

\bibitem[{Malik et~al.(2021)Malik, Sanjay, Nigam, Ghosh, Guha, Bhattacharya,
  and Modi}]{malik2021ildc}
Vijit Malik, Rishabh Sanjay, Shubham~Kumar Nigam, Kripabandhu Ghosh,
  Shouvik~Kumar Guha, Arnab Bhattacharya, and Ashutosh Modi. 2021.
\newblock Ildc for cjpe: Indian legal documents corpus for court judgment
  prediction and explanation.
\newblock In \emph{Proceedings of ACL-IJCNLP 2021}, pages 4046--4062.

\bibitem[{Martin et~al.(2020)Martin, Muller, Su{\'a}rez, Dupont, Romary,
  de~La~Clergerie, Seddah, and Sagot}]{martin2020camembert}
Louis Martin, Benjamin Muller, Pedro Javier~Ortiz Su{\'a}rez, Yoann Dupont,
  Laurent Romary, {\'E}ric~Villemonte de~La~Clergerie, Djam{\'e} Seddah, and
  Beno{\^\i}t Sagot. 2020.
\newblock Camembert: a tasty french language model.
\newblock In \emph{ACL 2020-58th Annual Meeting of the Association for
  Computational Linguistics}.

\bibitem[{Medvedeva et~al.(2021)Medvedeva, {\"U}st{\"u}n, Xu, Vols, and
  Wieling}]{medvedeva2021automatic}
Masha Medvedeva, Ahmet {\"U}st{\"u}n, Xiao Xu, Michel Vols, and Martijn
  Wieling. 2021.
\newblock Automatic judgement forecasting for pending applications of the
  european court of human rights.
\newblock In \emph{ASAIL/LegalAIIA@ ICAIL}.

\bibitem[{Medvedeva et~al.(2018)Medvedeva, Vols, and
  Wieling}]{medvedeva2018judicial}
Masha Medvedeva, Michel Vols, and Martijn Wieling. 2018.
\newblock Judicial decisions of the european court of human rights: Looking
  into the crystal ball.
\newblock In \emph{Proceedings of the conference on empirical legal studies},
  page~24.

\bibitem[{Mumcuo{\u{g}}lu et~al.(2021)Mumcuo{\u{g}}lu, {\"O}zt{\"u}rk, Ozaktas,
  and Ko{\c{c}}}]{mumcuouglu2021natural}
Emre Mumcuo{\u{g}}lu, Ceyhun~E {\"O}zt{\"u}rk, Haldun~M Ozaktas, and Aykut
  Ko{\c{c}}. 2021.
\newblock Natural language processing in law: Prediction of outcomes in the
  higher courts of turkey.
\newblock \emph{Information Processing \& Management}, 58(5):102684.

\bibitem[{Niklaus et~al.(2021)Niklaus, Chalkidis, and
  St{\"u}rmer}]{niklaus2021swiss}
Joel Niklaus, Ilias Chalkidis, and Matthias St{\"u}rmer. 2021.
\newblock Swiss-judgment-prediction: A multilingual legal judgment prediction
  benchmark.
\newblock In \emph{Proceedings of the Natural Legal Language Processing
  Workshop 2021}, pages 19--35.

\bibitem[{Niklaus and Giofr{\'e}(2022)}]{niklaus2022budgetlongformer}
Joel Niklaus and Daniele Giofr{\'e}. 2022.
\newblock Budgetlongformer: Can we cheaply pretrain a sota legal language model
  from scratch?
\newblock \emph{arXiv preprint arXiv:2211.17135}.

\bibitem[{Niklaus et~al.(2023{\natexlab{a}})Niklaus, Matoshi, Rani, Galassi,
  St{\"u}rmer, and Chalkidis}]{niklaus2023lextreme}
Joel Niklaus, Veton Matoshi, Pooja Rani, Andrea Galassi, Matthias St{\"u}rmer,
  and Ilias Chalkidis. 2023{\natexlab{a}}.
\newblock Lextreme: A multi-lingual and multi-task benchmark for the legal
  domain.
\newblock \emph{arXiv preprint arXiv:2301.13126}.

\bibitem[{Niklaus et~al.(2023{\natexlab{b}})Niklaus, Matoshi, Stürmer,
  Chalkidis, and Ho}]{niklaus2023multilegalpile}
Joel Niklaus, Veton Matoshi, Matthias Stürmer, Ilias Chalkidis, and Daniel~E.
  Ho. 2023{\natexlab{b}}.
\newblock \href {http://arxiv.org/abs/2306.02069} {Multilegalpile: A 689gb
  multilingual legal corpus}.

\bibitem[{Niklaus et~al.(2022)Niklaus, St{\"u}rmer, and
  Chalkidis}]{niklaus2022empirical}
Joel Niklaus, Matthias St{\"u}rmer, and Ilias Chalkidis. 2022.
\newblock An empirical study on cross-x transfer for legal judgment prediction.
\newblock In \emph{Proceedings of the 2nd Conference of the Asia-Pacific
  Chapter of the Association for Computational Linguistics and the 12th
  International Joint Conference on Natural Language Processing}, pages 32--46.

\bibitem[{Papineni et~al.(2002)Papineni, Roukos, Ward, and
  Zhu}]{papineni2002bleu}
Kishore Papineni, Salim Roukos, Todd Ward, and Wei-Jing Zhu. 2002.
\newblock Bleu: a method for automatic evaluation of machine translation.
\newblock In \emph{Proceedings of the 40th annual meeting of the Association
  for Computational Linguistics}, pages 311--318.

\bibitem[{Parisi et~al.(2020)Parisi, Francia, and Magnani}]{parisi2020umberto}
Loreto Parisi, Simone Francia, and Paolo Magnani. 2020.
\newblock Umberto: an italian language model trained with whole word masking.
\newblock \emph{Original-date}, 55:31Z.

\bibitem[{Rasiah et~al.(2023)Rasiah, Stern, Matoshi, Stürmer, Chalkidis, Ho,
  and Niklaus}]{rasiah_scale_2023}
Vishvaksenan Rasiah, Ronja Stern, Veton Matoshi, Matthias Stürmer, Ilias
  Chalkidis, Daniel~E. Ho, and Joel Niklaus. 2023.
\newblock \href {https://doi.org/10.48550/arXiv.2306.09237} {{SCALE}: {Scaling}
  up the {Complexity} for {Advanced} {Language} {Model} {Evaluation}}.
\newblock ArXiv:2306.09237 [cs].

\bibitem[{Read et~al.(2012)Read, Dridan, Oepen, and Solberg}]{read2012sentence}
Jonathon Read, Rebecca Dridan, Stephan Oepen, and Lars~J{\o}rgen Solberg. 2012.
\newblock Sentence boundary detection: A long solved problem?
\newblock In \emph{Proceedings of COLING 2012: Posters}, pages 985--994.

\bibitem[{Ribeiro et~al.(2016)Ribeiro, Singh, and Guestrin}]{ribeiro2016should}
Marco~Tulio Ribeiro, Sameer Singh, and Carlos Guestrin. 2016.
\newblock " why should i trust you?" explaining the predictions of any
  classifier.
\newblock In \emph{Proceedings of the 22nd ACM SIGKDD international conference
  on knowledge discovery and data mining}, pages 1135--1144.

\bibitem[{Ribeiro et~al.(2018)Ribeiro, Singh, and
  Guestrin}]{ribeiro2018anchors}
Marco~Tulio Ribeiro, Sameer Singh, and Carlos Guestrin. 2018.
\newblock Anchors: High-precision model-agnostic explanations.
\newblock In \emph{Proceedings of the AAAI conference on artificial
  intelligence}, volume~32.

\bibitem[{Rissland and Ashley(1987)}]{rissland1987case}
Edwina~L Rissland and Kevin~D Ashley. 1987.
\newblock A case-based system for trade secrets law.
\newblock In \emph{Proceedings of the 1st international conference on
  Artificial intelligence and law}, pages 60--66.

\bibitem[{Santosh et~al.(2023{\natexlab{a}})Santosh, Blas, Kemper, and
  Grabmair}]{santosh2023leveraging}
T.~Y. S.~S Santosh, Marcel Perez~San Blas, Phillip Kemper, and Matthias
  Grabmair. 2023{\natexlab{a}}.
\newblock Leveraging task dependency and contrastive learning for case outcome
  classification on european court of human rights cases.
\newblock \emph{arXiv preprint arXiv:2302.00768}.

\bibitem[{Santosh et~al.(2023{\natexlab{b}})Santosh, Ichim, and
  Grabmair}]{santosh2023Zero}
T.~Y. S.~S Santosh, Oana Ichim, and Matthias Grabmair. 2023{\natexlab{b}}.
\newblock Zero shot transfer of article-aware legal outcome classification for
  european court of human rights cases.
\newblock \emph{arXiv preprint arXiv:2302.00609}.

\bibitem[{Santosh et~al.(2022)Santosh, Xu, Ichim, and
  Grabmair}]{santosh2022deconfounding}
T.y.s.s Santosh, Shanshan Xu, Oana Ichim, and Matthias Grabmair. 2022.
\newblock Deconfounding legal judgment prediction for {E}uropean court of human
  rights cases towards better alignment with experts.
\newblock In \emph{Proceedings of EMNLP 2022}.

\bibitem[{Savelka et~al.(2017)Savelka, Walker, Grabmair, and
  Ashley}]{savelka2017sentence}
Jaromir Savelka, Vern~R Walker, Matthias Grabmair, and Kevin~D Ashley. 2017.
\newblock Sentence boundary detection in adjudicatory decisions in the united
  states.
\newblock \emph{Traitement automatique des langues}, 58:21.

\bibitem[{Semo et~al.(2022)Semo, Bernsohn, Hagag, Hayat, and
  Niklaus}]{semo2022classactionprediction}
Gil Semo, Dor Bernsohn, Ben Hagag, Gila Hayat, and Joel Niklaus. 2022.
\newblock Classactionprediction: A challenging benchmark for legal judgment
  prediction of class action cases in the us.
\newblock \emph{arXiv preprint arXiv:2211.00582}.

\bibitem[{Sert et~al.(2021)Sert, Y{\i}ld{\i}r{\i}m, and
  Ha{\c{s}}lak}]{sert2021using}
Mehmet~Fatih Sert, Engin Y{\i}ld{\i}r{\i}m, and {\.I}rfan Ha{\c{s}}lak. 2021.
\newblock Using artificial intelligence to predict decisions of the turkish
  constitutional court.
\newblock \emph{Social Science Computer Review}, page 08944393211010398.

\bibitem[{Shaikh et~al.(2020)Shaikh, Sahu, and Anand}]{shaikh2020predicting}
Rafe~Athar Shaikh, Tirath~Prasad Sahu, and Veena Anand. 2020.
\newblock Predicting outcomes of legal cases based on legal factors using
  classifiers.
\newblock \emph{Procedia Computer Science}, 167:2393--2402.

\bibitem[{Sharifi-Malvajerdi et~al.(2019)Sharifi-Malvajerdi, Kearns, and
  Roth}]{sharifi2019average}
Saeed Sharifi-Malvajerdi, Michael Kearns, and Aaron Roth. 2019.
\newblock Average individual fairness: Algorithms, generalization and
  experiments.
\newblock \emph{Advances in neural information processing systems}, 32.

\bibitem[{Strickson and De~La~Iglesia(2020)}]{strickson2020legal}
Benjamin Strickson and Beatriz De~La~Iglesia. 2020.
\newblock Legal judgement prediction for uk courts.
\newblock In \emph{Proceedings of the 2020 the 3rd international conference on
  information science and system}, pages 204--209.

\bibitem[{{\c{S}}ulea et~al.(2017{\natexlab{a}}){\c{S}}ulea, Zampieri, Malmasi,
  Vela, Dinu, and van Genabith}]{csulea2017exploring}
Octavia-Maria {\c{S}}ulea, Marcos Zampieri, Shervin Malmasi, Mihaela Vela,
  Liviu~P Dinu, and Josef van Genabith. 2017{\natexlab{a}}.
\newblock Exploring the use of text classification in the legal domain.

\bibitem[{{\c{S}}ulea et~al.(2017{\natexlab{b}}){\c{S}}ulea, Zampieri, Vela,
  and van Genabith}]{csulea2017predicting}
Octavia-Maria {\c{S}}ulea, Marcos Zampieri, Mihaela Vela, and Josef van
  Genabith. 2017{\natexlab{b}}.
\newblock Predicting the law area and decisions of french supreme court cases.
\newblock In \emph{Proceedings of the International Conference Recent Advances
  in Natural Language Processing, RANLP 2017}, pages 716--722.

\bibitem[{Sundararajan et~al.(2017)Sundararajan, Taly, and
  Yan}]{sundararajan2017axiomatic}
Mukund Sundararajan, Ankur Taly, and Qiqi Yan. 2017.
\newblock Axiomatic attribution for deep networks.
\newblock In \emph{ICML}, pages 3319--3328. PMLR.

\bibitem[{Tsarapatsanis and Aletras(2021)}]{tsarapatsanis2021ethical}
Dimitrios Tsarapatsanis and Nikolaos Aletras. 2021.
\newblock On the ethical limits of natural language processing on legal text.
\newblock In \emph{Findings of the Association for Computational Linguistics:
  ACL-IJCNLP 2021}, pages 3590--3599.

\bibitem[{Virtucio et~al.(2018)Virtucio, Aborot, Abonita, Avinante, Copino,
  Neverida, Osiana, Peramo, Syjuco, and Tan}]{virtucio2018predicting}
Michael Benedict~L Virtucio, Jeffrey~A Aborot, John Kevin~C Abonita, Roxanne~S
  Avinante, Rother Jay~B Copino, Michelle~P Neverida, Vanesa~O Osiana, Elmer~C
  Peramo, Joanna~G Syjuco, and Glenn Brian~A Tan. 2018.
\newblock Predicting decisions of the philippine supreme court using natural
  language processing and machine learning.
\newblock In \emph{2018 IEEE 42nd annual computer software and applications
  conference (COMPSAC)}, volume~2, pages 130--135. IEEE.

\bibitem[{Waltl et~al.(2017)Waltl, Bonczek, Scepankova, Landthaler, and
  Matthes}]{waltl2017predicting}
Bernhard Waltl, Georg Bonczek, Elena Scepankova, J{\"o}rg Landthaler, and
  Florian Matthes. 2017.
\newblock Predicting the outcome of appeal decisions in germany’s tax law.
\newblock In \emph{International conference on electronic participation}, pages
  89--99. Springer.

\bibitem[{Wang et~al.(2021)Wang, Xiao, Ma, Zhong, Tu, Zhang, Liu, and
  Sun}]{wang2021equality}
Yuzhong Wang, Chaojun Xiao, Shirong Ma, Haoxi Zhong, Cunchao Tu, Tianyang
  Zhang, Zhiyuan Liu, and Maosong Sun. 2021.
\newblock Equality before the law: Legal judgment consistency analysis for
  fairness.
\newblock \emph{arXiv preprint arXiv:2103.13868}.

\bibitem[{Wolf et~al.(2020)Wolf, Debut, Sanh, Chaumond, Delangue, Moi, Cistac,
  Rault, Louf, Funtowicz et~al.}]{wolf2020transformers}
Thomas Wolf, Lysandre Debut, Victor Sanh, Julien Chaumond, Clement Delangue,
  Anthony Moi, Pierric Cistac, Tim Rault, R{\'e}mi Louf, Morgan Funtowicz,
  et~al. 2020.
\newblock Transformers: State-of-the-art natural language processing.
\newblock In \emph{Proceedings of the 2020 conference on empirical methods in
  natural language processing: system demonstrations}, pages 38--45.

\bibitem[{Wu et~al.(2019)Wu, Zhang, and Wu}]{wu2019counterfactual}
Yongkai Wu, Lu~Zhang, and Xintao Wu. 2019.
\newblock Counterfactual fairness: Unidentification, bound and algorithm.
\newblock In \emph{Proceedings of the Twenty-Eighth International Joint
  Conference on Artificial Intelligence}.

\bibitem[{Xu et~al.(2023)Xu, T.y.s.s, Ichim, Risini, Plank, and
  Grabmair}]{xu2023dissonance}
Shanshan Xu, Santosh T.y.s.s, Oana Ichim, Isabella Risini, Barbara Plank, and
  Matthias Grabmair. 2023.
\newblock From dissonance to insights: Dissecting disagreements in rationale
  construction for case outcome classification.
\newblock In \emph{Proceedings of the 2023 Conference on Empirical Methods in
  Natural Language Processing}, pages 9558--9576, Singapore. Association for
  Computational Linguistics.

\bibitem[{Yue et~al.(2021)Yue, Liu, Jin, Wu, Zhang, An, Cheng, Yin, and
  Wu}]{yue2021neurjudge}
Linan Yue, Qi~Liu, Binbin Jin, Han Wu, Kai Zhang, Yanqing An, Mingyue Cheng,
  Biao Yin, and Dayong Wu. 2021.
\newblock Neurjudge: a circumstance-aware neural framework for legal judgment
  prediction.
\newblock In \emph{Proceedings of the 44th International ACM SIGIR Conference
  on Research and Development in Information Retrieval}, pages 973--982.

\bibitem[{Yurochkin et~al.()Yurochkin, Bower, and Sun}]{yurochkintraining}
Mikhail Yurochkin, Amanda Bower, and Yuekai Sun.
\newblock Training individually fair ml models with sensitive subspace
  robustness.
\newblock In \emph{International Conference on Learning Representations}.

\bibitem[{Zafar et~al.(2017)Zafar, Valera, Gomez~Rodriguez, and
  Gummadi}]{zafar2017fairness}
Muhammad~Bilal Zafar, Isabel Valera, Manuel Gomez~Rodriguez, and Krishna~P
  Gummadi. 2017.
\newblock Fairness beyond disparate treatment \& disparate impact: Learning
  classification without disparate mistreatment.
\newblock In \emph{Proceedings of the 26th international conference on world
  wide web}, pages 1171--1180.

\bibitem[{Zeiler and Fergus(2014)}]{zeiler2014visualizing}
Matthew~D Zeiler and Rob Fergus. 2014.
\newblock Visualizing and understanding convolutional networks.
\newblock In \emph{Computer Vision--ECCV 2014: 13th European Conference,
  Zurich, Switzerland, September 6-12, 2014, Proceedings, Part I 13}, pages
  818--833. Springer.

\bibitem[{Zhang and Bareinboim(2018)}]{zhang2018fairness}
Junzhe Zhang and Elias Bareinboim. 2018.
\newblock Fairness in decision-making—the causal explanation formula.
\newblock In \emph{Proceedings of the AAAI Conference on Artificial
  Intelligence}, volume~32.

\bibitem[{Zhang et~al.(2019)Zhang, Kishore, Wu, Weinberger, and
  Artzi}]{zhang2019bertscore}
Tianyi Zhang, Varsha Kishore, Felix Wu, Kilian~Q Weinberger, and Yoav Artzi.
  2019.
\newblock Bertscore: Evaluating text generation with bert.
\newblock In \emph{International Conference on Learning Representations}.

\bibitem[{Zhong et~al.(2020)Zhong, Wang, Tu, Zhang, Liu, and
  Sun}]{zhong2020iteratively}
Haoxi Zhong, Yuzhong Wang, Cunchao Tu, Tianyang Zhang, Zhiyuan Liu, and Maosong
  Sun. 2020.
\newblock Iteratively questioning and answering for interpretable legal
  judgment prediction.
\newblock In \emph{Proceedings of the AAAI Conference on Artificial
  Intelligence}, volume~34, pages 1250--1257.

\end{thebibliography}


\appendix
\section{Appendix}

\subsection{Explainability performance for different levels of occlusion}
We report the label wise F1-score for each occlusion test for every language in Tables \ref{occ_de}, \ref{occ_fr}, \ref{occ_it}. Overall, "Opposes judgement" and "Neutral" are challenging ones compared to "Supports judgement". In the case of French, there was an improvement in scores as the number of occluded sentences increased. This improvement indicates that the model was able to correctly associate the label "supports judgements" with the occluded sentences, thereby enhancing the model's explainability performance. However, in the case of French and Italian, a different trend was observed. The model did not exhibit the same improvement as the number of occluded sentences increased. It is speculated that the model might have encountered conflicting labels for each occluded sentence, leading to incorrect predictions when multiple occlusions were present.
\begin{table*}[]
\scalebox{0.65}{
\begin{tabular}
{|l|ccc|ccc|ccc|ccc|}
\toprule
                  & \multicolumn{3}{c|}{\textbf{Set 1}}                                                               & \multicolumn{3}{c|}{\textbf{Set 2}}                                                               & \multicolumn{3}{c|}{\textbf{Set 3}}                                                               & \multicolumn{3}{c|}{\textbf{Set 4}}                                                               \\ 
Model             & \multicolumn{1}{c|}{\textbf{Opposes}} & \multicolumn{1}{c|}{\textbf{Neutral}} & \textbf{Supports} & \multicolumn{1}{c|}{\textbf{Opposes}} & \multicolumn{1}{c|}{\textbf{Neutral}} & \textbf{Supports} & \multicolumn{1}{c|}{\textbf{Opposes}} & \multicolumn{1}{c|}{\textbf{Neutral}} & \textbf{Supports} & \multicolumn{1}{c|}{\textbf{Opposes}} & \multicolumn{1}{c|}{\textbf{Neutral}} & \textbf{Supports} \\ \midrule
MonoLingual       & \multicolumn{1}{c|}{23.08}            & \multicolumn{1}{c|}{20.98}            & 33.86             & \multicolumn{1}{c|}{8.80}             & \multicolumn{1}{c|}{17.85}            & 27.30             & \multicolumn{1}{c|}{3.28}             & \multicolumn{1}{c|}{20.19}            & 16.84             & \multicolumn{1}{c|}{1.15}             & \multicolumn{1}{c|}{21.34}            & 11.89             \\ \midrule
MultiLingual      & \multicolumn{1}{c|}{23.79}            & \multicolumn{1}{c|}{18.05}            & 34.46             & \multicolumn{1}{c|}{8.12}             & \multicolumn{1}{c|}{13.87}            & 26.81             & \multicolumn{1}{c|}{1.52}             & \multicolumn{1}{c|}{12.79}            & 24.25             & \multicolumn{1}{c|}{1.08}             & \multicolumn{1}{c|}{14..10}           & 11.17             \\ \midrule
MonoLingual + DA  & \multicolumn{1}{c|}{24.05}            & \multicolumn{1}{c|}{21.77}            & 25.12             & \multicolumn{1}{c|}{10.69}            & \multicolumn{1}{c|}{19.29}            & 17.34             & \multicolumn{1}{c|}{3.93}             & \multicolumn{1}{c|}{18.24}            & 10.01             & \multicolumn{1}{c|}{0.76}             & \multicolumn{1}{c|}{10.80}            & 22.03             \\ \midrule
MultiLingual + DA & \multicolumn{1}{c|}{28.80}            & \multicolumn{1}{c|}{26.40}            & 36.44             & \multicolumn{1}{c|}{12.45}            & \multicolumn{1}{c|}{21.64}            & 27.98             & \multicolumn{1}{c|}{4.49}             & \multicolumn{1}{c|}{19.83}            & 17.15             & \multicolumn{1}{c|}{1.04}             & \multicolumn{1}{c|}{18.01}            & 11.08             \\ \midrule
Joint Training      & \multicolumn{1}{c|}{24.52}            & \multicolumn{1}{c|}{20.14}            & 31.87             & \multicolumn{1}{c|}{9.86}             & \multicolumn{1}{c|}{10.62}            & 28.25             & \multicolumn{1}{c|}{2.69}             & \multicolumn{1}{c|}{9.20}             & 27.52             & \multicolumn{1}{c|}{0.98}             & \multicolumn{1}{c|}{5.75}             & 26.14             \\ \midrule
Joint Training + DA & \multicolumn{1}{c|}{23.85}            & \multicolumn{1}{c|}{23.33}            & 38.3              & \multicolumn{1}{c|}{13.40}            & \multicolumn{1}{c|}{16.02}            & 25.11             & \multicolumn{1}{c|}{4.71}             & \multicolumn{1}{c|}{10.74}            & 12.67             & \multicolumn{1}{c|}{1.24}             & \multicolumn{1}{c|}{7.77}             & 8.74              \\ \bottomrule
\end{tabular}}
\caption{Explainability performance for German dataset over different occlusion test sets}
\label{occ_de}
\end{table*}
\begin{table*}[]
\scalebox{0.65}{
\begin{tabular}{|l|ccc|ccc|ccc|ccc|}
\toprule
                  & \multicolumn{3}{c|}{\textbf{Set 1}}                                                               & \multicolumn{3}{c|}{\textbf{Set 2}}                                                               & \multicolumn{3}{c|}{\textbf{Set 3}}                                                               & \multicolumn{3}{c|}{\textbf{Set 4}}                                                               \\ 
Model             & \multicolumn{1}{c|}{\textbf{Opposes}} & \multicolumn{1}{c|}{\textbf{Neutral}} & \textbf{Supports} & \multicolumn{1}{c|}{\textbf{Opposes}} & \multicolumn{1}{c|}{\textbf{Neutral}} & \textbf{Supports} & \multicolumn{1}{c|}{\textbf{Opposes}} & \multicolumn{1}{c|}{\textbf{Neutral}} & \textbf{Supports} & \multicolumn{1}{c|}{\textbf{Opposes}} & \multicolumn{1}{c|}{\textbf{Neutral}} & \textbf{Supports} \\ \midrule
MonoLingual       & \multicolumn{1}{c|}{21.48}            & \multicolumn{1}{c|}{10.17}            & 36.67             & \multicolumn{1}{c|}{4.34}             & \multicolumn{1}{c|}{4.87}             & 38.17             & \multicolumn{1}{c|}{1.34}             & \multicolumn{1}{c|}{4.17}             & 53.64             & \multicolumn{1}{c|}{0.18}             & \multicolumn{1}{c|}{3.58}             & 67.99             \\ \midrule
MultiLingual      & \multicolumn{1}{c|}{26.80}            & \multicolumn{1}{c|}{2.40}             & 40.98             & \multicolumn{1}{c|}{6.19}             & \multicolumn{1}{c|}{0.68}             & 38.14             & \multicolumn{1}{c|}{1.05}             & \multicolumn{1}{c|}{0.79}             & 40.19             & \multicolumn{1}{c|}{0.00}             & \multicolumn{1}{c|}{0.22}             & 51.58             \\ \midrule
MonoLingual + DA  & \multicolumn{1}{c|}{20.05}            & \multicolumn{1}{c|}{10.53}            & 28.57             & \multicolumn{1}{c|}{4.72}             & \multicolumn{1}{c|}{8.67}             & 43.29             & \multicolumn{1}{c|}{1.28}             & \multicolumn{1}{c|}{6.79}             & 58.81             & \multicolumn{1}{c|}{0.14}             & \multicolumn{1}{c|}{4.35}             & 69.36             \\ \midrule
MultiLingual + DA & \multicolumn{1}{c|}{21.11}            & \multicolumn{1}{c|}{4.73}             & 30.41             & \multicolumn{1}{c|}{6.17}             & \multicolumn{1}{c|}{2.02}             & 38.24             & \multicolumn{1}{c|}{1.03}             & \multicolumn{1}{c|}{1.26}             & 41.39             & \multicolumn{1}{c|}{0.00}             & \multicolumn{1}{c|}{1.87}             & 40.86             \\ \midrule
Joint Training      & \multicolumn{1}{c|}{21.43}            & \multicolumn{1}{c|}{3.49}             & 37.37             & \multicolumn{1}{c|}{7.09}             & \multicolumn{1}{c|}{0.68}             & 43.02             & \multicolumn{1}{c|}{1.11}             & \multicolumn{1}{c|}{1.12}             & 49.77             & \multicolumn{1}{c|}{0.12}             & \multicolumn{1}{c|}{1.12}             & 58.29             \\ \midrule
Joint Training + DA & \multicolumn{1}{c|}{28.11}            & \multicolumn{1}{c|}{14.05}            & 38.78             & \multicolumn{1}{c|}{8.51}             & \multicolumn{1}{c|}{9.69}             & 44.26             & \multicolumn{1}{c|}{1.21}             & \multicolumn{1}{c|}{10.09}            & 43.65             & \multicolumn{1}{c|}{0.00}             & \multicolumn{1}{c|}{8.80}             & 42.04             \\ \bottomrule
\end{tabular}}
\caption{Explainability performance for French dataset over different occlusion test sets}
\label{occ_fr}
\end{table*}
\begin{table*}[]
\scalebox{0.65}{
\begin{tabular}{|l|ccc|ccc|ccc|ccc|}
\toprule
                  & \multicolumn{3}{c|}{\textbf{Set 1}}                                                               & \multicolumn{3}{c|}{\textbf{Set 2}}                                                               & \multicolumn{3}{c|}{\textbf{Set 3}}                                                               & \multicolumn{3}{c|}{\textbf{Set 4}}                                                               \\ 
Model             & \multicolumn{1}{c|}{\textbf{Opposes}} & \multicolumn{1}{c|}{\textbf{Neutral}} & \textbf{Supports} & \multicolumn{1}{c|}{\textbf{Opposes}} & \multicolumn{1}{c|}{\textbf{Neutral}} & \textbf{Supports} & \multicolumn{1}{c|}{\textbf{Opposes}} & \multicolumn{1}{c|}{\textbf{Neutral}} & \textbf{Supports} & \multicolumn{1}{c|}{\textbf{Opposes}} & \multicolumn{1}{c|}{\textbf{Neutral}} & \textbf{Supports} \\ \midrule
MonoLingual       & \multicolumn{1}{c|}{9.01}             & \multicolumn{1}{c|}{24.70}            & 23.53             & \multicolumn{1}{c|}{2.01}             & \multicolumn{1}{c|}{26.56}            & 6.72              & \multicolumn{1}{c|}{0.52}             & \multicolumn{1}{c|}{31.08}            & 0.22              & \multicolumn{1}{c|}{0.08}             & \multicolumn{1}{c|}{32.43}            & 0.00              \\ \midrule
MultiLingual      & \multicolumn{1}{c|}{12.84}            & \multicolumn{1}{c|}{19.18}            & 31.25             & \multicolumn{1}{c|}{2.54}             & \multicolumn{1}{c|}{7.66}             & 14.86             & \multicolumn{1}{c|}{0.41}             & \multicolumn{1}{c|}{7.22}             & 4.84              & \multicolumn{1}{c|}{0.0}              & \multicolumn{1}{c|}{3.92}             & 1.08              \\ \midrule
MonoLingual + DA  & \multicolumn{1}{c|}{22.11}            & \multicolumn{1}{c|}{3.03}             & 27.10             & \multicolumn{1}{c|}{5.61}             & \multicolumn{1}{c|}{0.24}             & 13.37             & \multicolumn{1}{c|}{0.84}             & \multicolumn{1}{c|}{0.33}             & 6.12              & \multicolumn{1}{c|}{0.05}             & \multicolumn{1}{c|}{0.35}             & 1.92              \\ \midrule
MultiLingual + DA & \multicolumn{1}{c|}{21.28}            & \multicolumn{1}{c|}{11.43}            & 35.82             & \multicolumn{1}{c|}{8.66}             & \multicolumn{1}{c|}{2.62}             & 20.77             & \multicolumn{1}{c|}{0.76}             & \multicolumn{1}{c|}{1.06}             & 8.51              & \multicolumn{1}{c|}{0.06}             & \multicolumn{1}{c|}{0.18}             & 2.46              \\ \midrule
Joint Training      & \multicolumn{1}{c|}{17.24}            & \multicolumn{1}{c|}{11.21}            & 25.61             & \multicolumn{1}{c|}{3.27}             & \multicolumn{1}{c|}{5.61}             & 17.55             & \multicolumn{1}{c|}{0.37}             & \multicolumn{1}{c|}{1.79}             & 8.07              & \multicolumn{1}{c|}{0.0}              & \multicolumn{1}{c|}{0.63}             & 3.01              \\ \midrule
Joint Training + DA & \multicolumn{1}{c|}{20.99}            & \multicolumn{1}{c|}{17.94}            & 33.53             & \multicolumn{1}{c|}{3.99}             & \multicolumn{1}{c|}{6.50}             & 23.02             & \multicolumn{1}{c|}{0.46}             & \multicolumn{1}{c|}{2.28}             & 11.3              & \multicolumn{1}{c|}{0.04}             & \multicolumn{1}{c|}{0.77}             & 3.91              \\ \bottomrule
\end{tabular}}
\caption{Explainability performance for Italian dataset over different occlusion test sets}
\label{occ_it}
\end{table*}

\end{document}